\documentclass{article}

\usepackage{spconf}
\usepackage[english]{babel}
\usepackage[utf8]{inputenc}
\usepackage{times} 
\usepackage{cite}

\usepackage{hyperref}
\usepackage[nolist]{acronym}

\usepackage{multirow}

\usepackage{siunitx}
\usepackage{graphicx} 
\usepackage{epsfig} 
\usepackage{pgfplots}
\usepackage{caption}
\usepackage[font=footnotesize]{subcaption}
\usepackage{pgfplots}

\usepackage{float}

\usepackage[export]{adjustbox}

\usepackage{mathtools}
\usepackage{nicefrac}
\usepackage{amsmath}
\usepackage{amssymb}
\usepackage{bm} 

\DeclareMathOperator*{\maximize}{maximize}

\usepackage{etoolbox}
\makeatletter%
\AfterPreamble{%
	\usepackage{hyperref}%
	\let\oldhypertarget\hypertarget%
	\renewcommand{\hypertarget}[2]{%
		\oldhypertarget{#1}{#2}%
		\protected@write\@mainaux{}{%
			\string\expandafter\string\gdef%
			\string\csname\string\detokenize{#1}\string\endcsname{#2}%
		}%
	}%
	\newcommand{\myhyperlink}[1]{%
		\hyperlink{#1}{\csname #1\endcsname}%
	}%
}
\makeatother%

\newcounter{Remark}
\newcounter{Problem}
\usepackage{algorithm}
\usepackage[noend]{algpseudocode}

\makeatletter
\def\BState{\State\hskip-\ALG@thistlm}
\makeatother

\usepackage{tikz}
\pgfdeclarelayer{foreground}
\pgfsetlayers{background, main, foreground}
\usetikzlibrary{quotes, angles, backgrounds, arrows, automata, shapes, positioning, calc, through, spy, decorations.markings}

\usepackage{aircraftshapes} 

\tikzset{
	imglabel/.style={
		rectangle,
		inner sep=2pt,
		text=black,
		minimum height=1em,
		text centered,
		fill=white,
		fill opacity=1.0,
		text opacity=1,
		anchor=south west,
	},
}

\tikzset{
	state/.style={
		rectangle,
		draw=black, very thick,
		minimum height=1.0em,
		text centered,
	},
}

\newcommand\copyrighttext{%
    \small \begin{center} \color{red} \textcopyright\,2024 IEEE. Personal use of this material is permitted. Permission from IEEE must be obtained for all other uses, in any current or future media, including reprinting/republishing this material for advertising or promotional purposes, creating new collective works, for resale or redistribution to servers or lists, or reuse of any copyrighted component of this work in other works. \end{center}}

\title{\vspace{-2em} \copyrighttext \vspace{1em} \large Omnidirectional Multi-Rotor Aerial Vehicle Pose Optimization: a novel approach to physical layer security}

\name{Daniel Bonilla Licea${^{1,2}}$\thanks{This work was partially funded by the EU's H2020 research and innovation programme AERIAL-CORE grant no. 871479, by the CTU grant no. SGS23/177/OHK3/3T/13, by the Czech Science Foundation (GAČR) grant no. 23-07517S, and by the EU under the project Robotics and advanced industrial production (reg. no. CZ.02.01.01/00/22\_008/0004590).}, Giuseppe Silano${^2}$, Mounir Ghogho${^3}$, and Martin Saska${^2}$}
\address{$^1$College of Computing, Mohammed VI Polytechnic University, Ben Guerir, Morocco,~\\~(email: {\tt\normalsize daniel.bonilla@um6p.ma})\\
$^2$Faculty of Electrical Engineering, Czech Technical University in Prague, Czech Republic, \\ 
(emails: {\tt\normalsize \{giuseppe.silano, martin.saska\}@fel.cvut.cz})\\
 $^3$International University of Rabat, Morocco, (email: {\tt\normalsize mounir.ghogho@uir.ac.ma})}

\begin{document}
%
\maketitle



\begin{acronym}
    \acro{AoA}[AoA]{Angle of Arrival}
    \acro{AoD}[AoD]{Angle of Departure}
    \acro{BS}[BS]{Base Station}
    \acro{DoF}[DoF]{Degree of Freedom}
    \acro{FDMA}[FDMA]{Frequency-Division Multiple Access}
    \acro{LoS}[LoS]{Line of Sight}
    \acro{MRAV}[MRAV]{Multi-Rotor Aerial Vehicle}
    \acro{UAV}[UAV]{Unmanned Aerial Vehicle} 
    \acro{SNR}[SNR]{Signal-to-Noise Ratio}
    \acro{SINR}[SINR]{Signal-to-Interference-plus-Noise Ratio}
    \acro{w.r.t.}[w.r.t.]{with respect to}
\end{acronym}



\begin{abstract}

The integration of~\acp{MRAV} into 5G and 6G networks enhances coverage, connectivity, and congestion management. This fosters communication-aware robotics, exploring the interplay between robotics and communications, but also makes the~\acp{MRAV} susceptible to malicious attacks, such as jamming. One traditional approach to counter these attacks is the use of beamforming on the~\acp{MRAV} to apply physical layer security techniques. In this paper, we explore \textit{pose optimization} as an alternative approach to countering jamming attacks on~\acp{MRAV}. This technique is intended for omnidirectional~\acp{MRAV}, which are drones capable of independently controlling both their position and orientation, as opposed to the more common under-actuated~\acp{MRAV} whose orientation cannot be controlled independently of their position. In this paper, we consider an omnidirectional~\ac{MRAV} serving as a~\ac{BS} for legitimate ground nodes, under attack by a malicious jammer. We optimize the~\ac{MRAV} pose (i.e., position and orientation) to maximize the minimum~\ac{SINR} over all legitimate nodes.

\end{abstract}






\section{Introduction}
\label{sec:introduction}

Recently, there has been a remarkable surge in the research field of communications-aware robotics, as demonstrated by the growing number of publications on this subject~\cite{Gasparri2017TRO, Licea2020TRO, Zeng2017TWC, Wu2018TWC}. The interest in exploring the connection between communications and robotics can be partially attributed to the ongoing advancements in 5G and forthcoming 6G technologies. These technologies aim to integrate~\acfp{MRAV} into the cellular communications network, with the goal of enhancing coverage and connectivity, bolstering network resilience, and alleviating congestion by offloading data traffic, among other benefits~\cite{Zeng2019ProceedingsIEEE, Muralidharan2021ARCRAS}. A significant portion of this research focuses on under-actuated~\acp{MRAV}~\cite{Jung2010CM, LindheICRA2010, CalvoFullana2021IEEECM}.

Under-actuated~\acp{MRAV}~\cite{Hamandi2021IJRR} exhibit the ability to hover at specific positions and can serve as aerial~\acfp{BS}~\cite{Licea2023ICUAS, Kishk2020VTM}. Additionally, they can also track trajectories, enabling them to function as mobile communications relays~\cite{Licea2021EUSIPCO, VaradharajanRAL2020}. But, one limitation of under-actuated~\acp{MRAV} is their inability to independently control both their position and orientation. This constraint arises from the fact that under-actuated~\acp{MRAV} typically have fewer control inputs (i.e., rotors or propellers) than the number of~\acp{DoF} needed for fully independent position and orientation control~\cite{Hamandi2021IJRR}. 
This means that the position and orientation of the fixed antenna mounted on the~\ac{MRAV} cannot be controlled independently. Additionally, the~\ac{SNR} depends not only on the~\ac{MRAV} position, but also on its tilt. Thus, this underactuation represents an obstacle in the maximization of the~\ac{SNR}.

\begin{figure}[tb]
    \centering
    \scalebox{0.925}{
    \begin{tikzpicture}
        \node at (0,0) [text centered] {\adjincludegraphics[trim={{.0\width} {.0\height} {.0\width} {.0\height}}, clip, width=0.9\columnwidth]{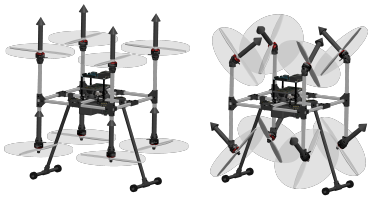}};

        \draw[-latex,red] (-1.85,0) -- (-1.85,1.5) node[above]{{\scriptsize{$z_U$}}}; 
        \draw[-latex,red] (-1.85,0) -- (-0.35,-0.21) node[right]{{\scriptsize{$y_U$}}}; 
        \draw[-latex,red] (-1.85,0) -- (-2.85,-0.31) node[below,red]{{\scriptsize{$x_U$}}}; 
        \node at (-1.85,0) [below]{{\scriptsize{$O_U$}}};

        \draw[-latex] (-3.85,-2) -- (-3.85,-1.5) node[above]{{\scriptsize{$z_W$}}}; 
        \draw[-latex] (-3.85,-2) -- (-3.25,-2.29) node[above]{{\scriptsize{$y_W$}}}; 
        \draw[-latex] (-3.85,-2) -- (-4.55,-2.19) node[above]{{\scriptsize{$x_W$}}}; 
        \node at (-3.85,-2) [below]{{\scriptsize{$O_W$}}};

        \draw[-latex, dashed,red] (-3.85,-2) -- node[right]{{\scriptsize{$\mathbf{p}_{BS}$}}}(-1.85,0);
    \end{tikzpicture}
    }
    \vspace*{-1em}
    \caption{Illustration of two~\ac{MRAV} configurations along with the global ($\mathcal{F}_W$) and untilted ($\mathcal{F}_U$) reference systems: under-actuated (left) and omnidirectional (right)~\cite{Aboudorra2023ArXiv}.}
    \label{fig:mrav}
\end{figure}
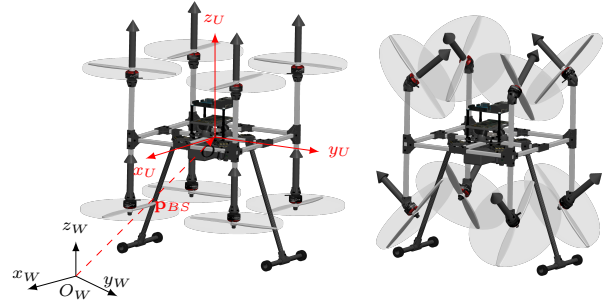

Conversely, omnidirectional~\acp{MRAV} offer the distinct advantage of simultaneous control over both their position and orientation~\cite{Allenspach2020IJRR, Aboudorra2023ArXiv}. Omnidirectionality refers to a vehicle's capacity to support its weight in any orientation, making it particularly advantageous for dealing with jamming attacks which have become an increasing threat to \acp{UAV} \cite{LeccaditoIEEEAESM2018,SedjelmaciIEEETSMCS2018}. An omnidirectional~\ac{MRAV} can adjust its orientation and precisely direct its antenna null towards the malicious jammer to neutralize it, while also maintaining favourable channel gain with legitimate communication nodes. Although similar effects can be achieved using beamforming techniques~\cite{Xiao2022CST}, this approach often requires an array of antennas, which may not be feasible due to size constraints, increased costs, and higher energy consumption for small vehicles~\cite{VADLAMANI2016IJPE}. In contrast, omnidirectional~\acp{MRAV} can achieve such similar outcomes without any additional hardware. Figure~\ref{fig:mrav} presents an illustrative example of both an under-actuated and an omnidirectional~\ac{MRAV}. 

However, harnessing the inherent omnidirectional capabilities of these platforms requires the utilization of advanced motion planning techniques. These techniques play a pivotal role in enabling these aerial vehicles to effectively position and orient themselves, thereby optimizing their communication performance while mitigating potential jamming. In this context, this paper presents an innovative method for calculating the pose (i.e., position and orientation) of an omnidirectional~\ac{MRAV} acting as an aerial~\acl{BS} to minimize the~\acf{SINR}. The considered scenario involves the~\ac{MRAV} receiving data from a set of $N$ stationary nodes, while contending with the presence of a stationary malicious node, denoted as $M$. Notably, the antenna is positioned on the upper surface of the \ac{MRAV}. 
To the best of the authors' knowledge, this work marks the first application of such a strategy within the domain of the physical layer security for drone communications.

\section{System Model}
\label{sec:systemModel}

Let us consider a legitimate communications network consisting of $N$ stationary nodes $\{{S}_i\}_{i=1}^N$ and an omnidirectional~\ac{MRAV} functioning as an aerial~\ac{BS}. Simultaneously, a stationary malicious node $M$ is positioned within the same operational area. We denote the positions of $\{{S}_i\}_{i=1}^N$, the \ac{MRAV}, and the malicious node $M$ in the global reference frame $\mathcal{F}_W = \{O_W, \mathbf{x}_W, \mathbf{y}_W, \mathbf{z}_W\}$ as $\mathbf{p}_{S_i}$, $\mathbf{p}_{BS}$, and $\mathbf{p}_{M}$, respectively. For convenience, we introduce the untilted coordinate frame $\mathcal{F}_U = \{O_U, \mathbf{x}_U, \mathbf{y}_U, \mathbf{z}_U\}$, aligned with the global coordinate frame $\mathcal{F}_W$, and centered at $\mathbf{p}_{BS}$ (see Figure~\ref{fig:mrav}). To precisely describe the orientation of the~\ac{MRAV} in the global coordinate frame, Euler angles are employed, specifically roll ($\varphi$), pitch ($\vartheta$), and yaw ($\psi$). We refer to the orientation of the~\ac{MRAV} as $\bm{\eta}_{BS}=[\varphi, \vartheta, \psi]^\top$.

We assume that the omnidirectional \ac{MRAV} is equipped with a single antenna, located at $\mathbf{p}_{BS}$, on its upper surface, and oriented according to the following vector, expressed in $\mathcal{F}_U$:
\begin{equation}\label{eq:1.1}
\bm{\Upsilon}(\bm{\eta}_{BS})= 
    \begin{bmatrix}
        \cos(\varphi)\sin(\vartheta)\cos(\psi)+\sin(\varphi)\sin(\psi)\\
        \cos(\varphi)\sin(\vartheta)\sin(\psi)-\sin(\varphi)\cos(\psi)\\
        \cos(\varphi)\cos(\vartheta)
   \end{bmatrix}.
\end{equation}
Without loss of generality, we consider this antenna to be a small dipole, and thus its normalized power radiation pattern can be  described as \cite{MIRON20069}:
\begin{equation}\label{eq:1.2}
    G(\gamma)=\sin^2(\gamma),
\end{equation}
where $\gamma$ represents the elevation angle component of the \ac{AoA}. Let us now focus on the communications link between the \ac{MRAV} and the node $S_i$. The cosine of the elevation angle component for this link is:
\begin{equation}\label{eq:1.3}
  \cos(\gamma_i) = \left\langle \frac{\mathbf{p}_{S_i}-\mathbf{p}_{BS}}{\|\mathbf{p}_{S_i}-\mathbf{p}_{BS}\|}, \bm{\Upsilon}(\bm{\eta}_{BS})\right\rangle,
\end{equation}
where $\langle\cdot,\cdot\rangle$ represents the inner product operation, and $\gamma_i$ is the \ac{AoA} of the signal emitted by node $S_i$. Subsequently, we can formulate the antenna channel gains for the signals received from the legitimate nodes as:
\begin{equation}\label{eq:1.4}
    G(\gamma_i) = 1-\left\langle \frac{\mathbf{p}_{S_i}-\mathbf{p}_{BS}}{\|\mathbf{p}_{S_i}-\mathbf{p}_{BS}\|}, \bm{\Upsilon}(\bm{\eta}_{BS})\right\rangle^2.
\end{equation}
For the signal received from the jammer, we replace $\gamma_i$ by $\gamma_M$ (the \ac{AoA} of the signal emitted by the malicious node $M$) and $\mathbf{p}_{S_i}$ by $\mathbf{p}_{M}$ in~\eqref{eq:1.4}. Lastly, for simplicity and to focus on the~\ac{MRAV}, we will assume that both $\{{S}_i\}_{i=1}^N$ and ${M}$ are equipped with isotropic antennas.



\section{Pose Optimization}
\label{sec:problemDescription}

We consider a scenario where 
the malicious node, ${M}$, acts as a jammer. 
The objective is to optimize the pose of the~\ac{MRAV} to maximize the minimum~\acl{SINR} (SINR) over all $N$ legitimate nodes.  

The nodes $\{{S}_i\}_{i=1}^N$ employ~\ac{FDMA} to transmit data to the~\ac{MRAV} over $N$ distinct frequency channels. We assume that all frequency channels are narrow and further consider the antenna's frequency bandwidth to be sufficiently large to ensure that the antenna gain is independent of the frequency channel. We also assume that there is no adjacent channel interference. As the legitimate nodes transmit data to the \ac{MRAV}, the malicious node $M$ emits a jamming signal characterized by a flat power spectral density that covers all communication channels, and all frequency channels experience uniform interference. 

For the sake of simplicity, we focus on the antenna radiation pattern and assume~\ac{LoS} conditions between the~\ac{MRAV} and all nodes (including both legitimate nodes and the malicious node). 
Hence, the \ac{SINR} encountered by the \ac{MRAV} in the link with the $i$-th node is given by:
\begin{equation}\label{eq:2.1}
    \Gamma_i = \frac{\left(\frac{G(\gamma_i)P}{\|\mathbf{p}_{S_i}-\mathbf{p}_{BS}\|^2}\right)}{\left(\frac{G(\gamma_M)P_M}{\|\mathbf{p}_M-\mathbf{p}_{BS}\|^2}\right)+\sigma^2}  
    %
    ,\; \text{with} \; i=\{1, 2, \dots, N\},
\end{equation}
where $P$ is the transmission power of the legitimate nodes, $P_M$ is the transmission power of the malicious node, and $\sigma^2$ is the power of the noise at the \ac{MRAV} receiver.

Our objective is to optimize the pose of the \ac{MRAV}, encompassing both its position and orientation, in order to maximize the minimum \ac{SINR} across all nodes. To achieve this goal, we formulate the following optimization problem:
\begin{subequations} \label{eq:2.2a}
    \begin{align}
    &\maximize_{\bm{\eta}_{BS}, \, \mathbf{p}_{BS}} \left( \min_{i} \; \Gamma_i \right) \label{subeq:2.2a_objective}\\
    &\quad \;\;\,\, \text{s.t.}~\quad \underline{z} \leq \mathbf{e}_3^\top \mathbf{p}_{BS} \leq \bar{z}, \label{subeq:2.2a_first}\\
    &\qquad \qquad \left\|
     \begin{bmatrix}
        \mathbf{I}_2 & \\
        & 0 \\
    \end{bmatrix} \bm{\eta}_{BS} \right\|_{\infty} \leq \frac{\pi}{2}, \label{subeq:2.2a_second} \\
    &\qquad \qquad \text{with} \; \psi=0. \label{subeq:2.2a_third}
    \end{align}
\end{subequations}

In this formulation, the objective function~\eqref{subeq:2.2a_objective} is designed to maximize the minimum~\ac{SINR} experienced by the set of $N$ stationary nodes. The constraint~\eqref{subeq:2.2a_first} limits the \ac{MRAV}'s altitude. Here, $\mathbf{e}_3$ denotes the third column of the identity matrix $\mathbf{I}_3 \in \mathbb{R}^{3 \times 3}$. The constraint~\eqref{subeq:2.2a_second} establishes acceptable ranges for the pitch ($\vartheta$) and roll ($\varphi$) angles. The search space of the optimization problem encompasses five dimensions ($\mathbf{p}_{BS}$, roll $\varphi$, and pitch $\vartheta$ angles). The omission of the yaw angle $\psi$ is based on the symmetry of the antenna radiation pattern around its axis. Notably, it is assumed that both transmission powers $P_M$ and $P$ are known to the \ac{MRAV}, as well as the positions of all the legitimate nodes and the position of the jammer\footnote{The position of the jammer can be estimated using techniques such as the one described in \cite{BhamidipatiIEEEIoTJ2019}.}. Additionally, due to the high nonlinearity of the optimization problem concerning the \ac{MRAV}'s pose, two suboptimal solutions are developed in this section. 



\subsection{Zero Interference}
\label{sec:Zero Interference}

Let us start by discussing the suboptimal solution called \textit{\it zero interference}. The idea behind this suboptimal solution is to always direct the null of the \ac{MRAV} antenna towards the malicious node $M$, effectively setting $G(\gamma_M)=0$. Consequently, the orientation of the \ac{MRAV} is adjusted to satisfy:
\begin{equation}\label{eq:3.1}
    \bm{\Upsilon}(\bm{\eta}_{BS})=  \frac{b(\mathbf{p}_{M}-\mathbf{p}_{BS})}{\|\mathbf{p}_{M}-\mathbf{p}_{BS}\|},  
\end{equation}
where $b=\{+1, -1\}$. As a result,~\eqref{eq:2.1} transforms into:
\begin{eqnarray}\label{eq:3.2}
    {\Gamma}_i^\mathrm{ZI}&=&\frac{\left(1-\left\langle \frac{\mathbf{p}_{S_i}-\mathbf{p}_{BS}}{\|\mathbf{p}_{S_i}-\mathbf{p}_{BS}\|}, \frac{b(\mathbf{p}_{M}-\mathbf{p}_{BS})}{\|\mathbf{p}_{M}-\mathbf{p}_{BS}\|} \right\rangle^2\right)P}{\|\mathbf{p}_{S_i}-\mathbf{p}_{BS}\|^2\sigma^2},  
\end{eqnarray}
with $i=\{1,2,\dots,N\}$. Consequently, problem~\eqref{eq:2.2a} can be reformulated as the following position optimization problem:
\begin{subequations} \label{eq:2.2b}
    \begin{align}
    &\maximize_{\mathbf{p}_{BS}} \left( \min_{i} \; \Gamma_i^\mathrm{ZI} \right) \label{subeq:2.2b_objective}\\
    &\quad \;\;\,\, \text{s.t.}~\quad \underline{z} \leq \mathbf{e}_3^\top \mathbf{p}_{BS} \leq \bar{z} \label{subeq:2.2b_first}.
    \end{align}
\end{subequations}



\subsection{Maximum Gain}
\label{sec:Max Gain}

Now we focus on the suboptimal solution called \textit{maximum gain}. The concept behind this suboptimal solution is to continuously direct the maximum antenna gain towards the legitimate nodes. In general, this can be accomplished only when $N=2$, and when the orientation of the \ac{MRAV} is:
\begin{equation}\label{eq:3.4}
    \bm{\Upsilon}(\bm{\eta}_{BS})=b\left(  \frac{\mathbf{p}_{S_1}-\mathbf{p}_{BS}}{\|\mathbf{p}_{S_1}-\mathbf{p}_{BS}\|}\right)\times \left(  \frac{\mathbf{p}_{S_2}-\mathbf{p}_{BS}}{\|\mathbf{p}_{S_2}-\mathbf{p}_{BS}\|}\right),  
\end{equation}
where $\times$ is the cross-product operator and $b=\{+1, -1\}$. Consequently,~\eqref{eq:2.1} transforms into:
\begin{equation} \label{eq:3.5}
    \resizebox{0.89\hsize}{!}{$%
    \Gamma_i^\mathrm{MG} = \frac{\frac{P}{\|\mathbf{p}_{S_i}-\mathbf{p}_{BS}\|^2}}{\frac{\left(1-\left\langle \frac{\mathbf{p}_{M}-\mathbf{p}_{BS}}{\|\mathbf{p}_{M}-\mathbf{p}_{BS}\|},   \frac{\mathbf{p}_{S_1}-\mathbf{p}_{BS}}{\|\mathbf{p}_{S_1}-\mathbf{p}_{BS}\|}\times  \frac{\mathbf{p}_{S_2}-\mathbf{p}_{BS}}{\|\mathbf{p}_{S_2}-\mathbf{p}_{BS}\|}  \right\rangle^2\right)P_M}{\|\mathbf{p}_{M}-\mathbf{p}_{BS}\|^2}+\sigma^2}.
    $}%
\end{equation}

With the orientation now fixed, the pose optimization problem~\eqref{eq:2.2a} transforms into a position optimization problem:
\begin{subequations} \label{eq:2.2c}
    \begin{align}
    &\maximize_{\mathbf{p}_{BS}} \left( \min_{i=\{1,2\}} \; \Gamma_i^\mathrm{MG} \right) \label{subeq:2.2c_objective}\\
    &\quad \;\;\,\, \text{s.t.}~\quad \underline{z} \leq \mathbf{e}_3^\top \mathbf{p}_{BS} \leq \bar{z} \label{subeq:2.2c_first}.
    \end{align}
\end{subequations}

The resultant optimization problem~\eqref{eq:2.2c} is nonconvex with multiple local optima. Therefore, we solve it numerically using methods like simulated annealing. Despite the complexity, we can still glean insights into the optimal position. Notably, $\Gamma_1^\mathrm{MG}$ and $\Gamma_2^\mathrm{MG}$ share the same denominator and differ only in their numerators. Consequently, we can express $\Gamma_1^\mathrm{MG}$ and $\Gamma_2^\mathrm{MG}$ as follows:
\begin{equation}
    \Gamma_i^\mathrm{MG}(\mathbf{p}_{BS})=\frac{\frac{P}{\|\mathbf{p}_{S_i}-\mathbf{p}_{BS}\|^2}}{D(\mathbf{p}_{BS})}=\frac{N_i(\mathbf{p}_{BS})}{D(\mathbf{p}_{BS})},
\end{equation}
with $i=\{1,2\}$. Let us assume $\mathbf{p}_{BS}^\star$ to be the optimal position for the problem~\eqref{eq:2.2c}, and assume that this optimum solution satisfies $\|\mathbf{p}_{S_2}-\mathbf{p}_{BS}^\star\| > \|\mathbf{p}_{S_1}-\mathbf{p}_{BS}^\star\|$. We assign the optimum value of the objective function as $J(\mathbf{p}_{BS}^\star) \triangleq \min(\Gamma_1^\mathrm{MG}(\mathbf{p}_{BS}^\star), \Gamma_2^\mathrm{MG}(\mathbf{p}_{BS}^\star))$. Now, consider a new position $\mathbf{p}_{BS}$ that results from slightly adjusting $\mathbf{p}_{BS}^\star$ to bring it closer to $\mathbf{p}_{S_2}$, while ensuring the following conditions are met:
\begin{eqnarray}
 D(\mathbf{p}_{BS})&=&D(\mathbf{p}_{BS}^\star),\\
 \|\mathbf{p}_{S_2}-\mathbf{p}_{BS}\|&>& \|\mathbf{p}_{S_1}-\mathbf{p}_{BS}\|,\\
 N_2(\mathbf{p}_{BS})&>& N_2(\mathbf{p}_{BS}^\star).
\end{eqnarray}
From this, we infer that $J(\mathbf{p}_{BS})>J(\mathbf{p}_{BS}^\star)$, implying that $\mathbf{p}_{BS}^\star$ is not optimal. Consequently, if the optimal position yields a denominator value $\tilde{D}(\mathbf{p}_{BS})$ and there exists a set of positions $\mathcal{P}$ with the same denominator value, then the optimal position minimizes the difference between $\|\mathbf{p}_{S_2}-\mathbf{p}_{BS}\|$ and $\|\mathbf{p}_{S_1}-\mathbf{p}_{BS}\|$.

Considering this analysis and extensive simulations across various conditions, we observed that the optimal position for the \textit{maximum gain} problem must be equidistant from both legitimate nodes. 


\section{Simulation Results}
\label{sec:simulations}

To gain deeper insight into the pose optimization technique discussed in this paper, we present numerical simulation results obtained using MATLAB. All numerical simulations were conducted on a computer equipped with an i7-8565U processor (1.80 GHz) and 32GB of RAM, running on the Ubuntu 20.04 operating system. 

We consider a scenario with two legitimate nodes, $N=2$, positioned as follows: $\mathbf{p}_{S_1}=[0 \;\; 0 \;\; 0]^\top$ and $\mathbf{p}_{S_2}=[0 \;\; 50 \;\; 0]^\top$. The altitude range is defined by $\underline{z}=8$ and $\bar{z}=30$.
The noise-to-transmission power ratio is $\sigma^2/P=0.001$. A malicious node is situated at $\mathbf{p}_{M}=[17 \;\; 15 \;\; 4]^\top$. The jamming-to-transmission power ratio $P_M/P$ varies.

\begin{figure}[tb]
    \centering
    \adjincludegraphics[trim={{.05\width} {.225\height} {.05\width} {.225\height}},clip,width=0.90\columnwidth]{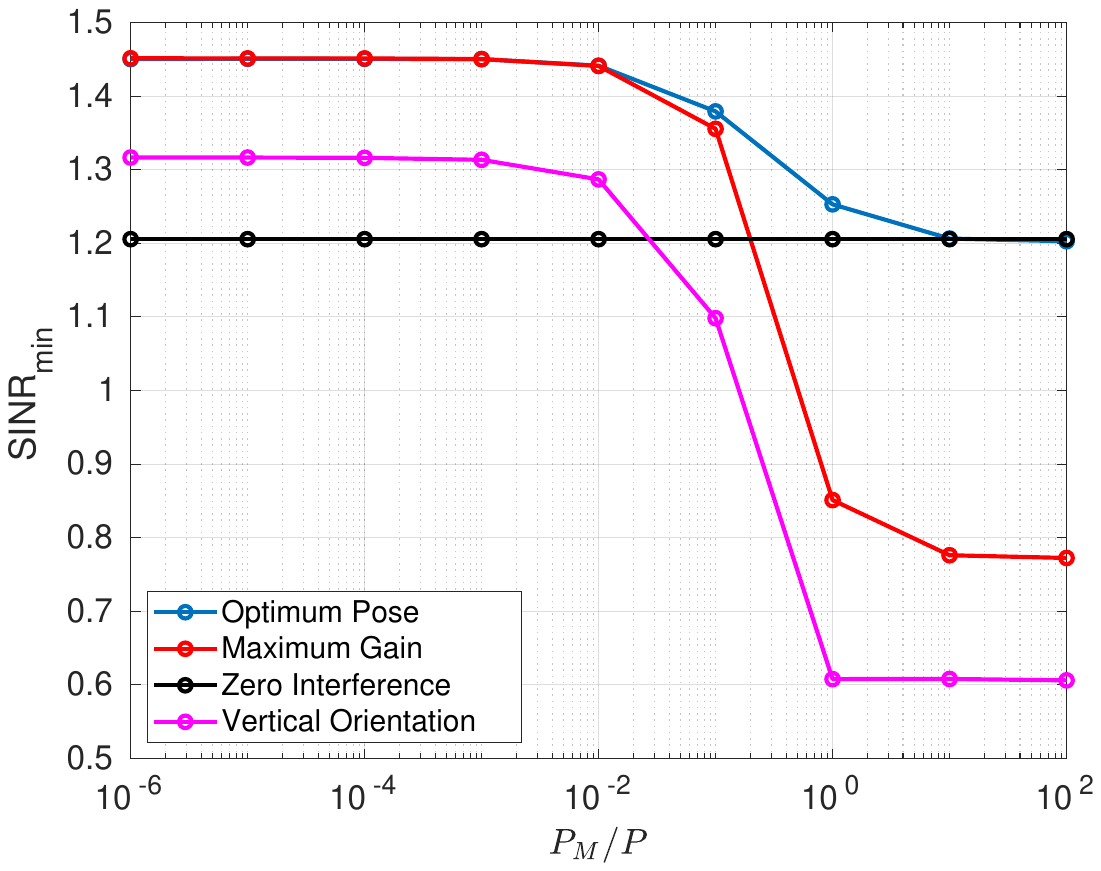}
    \vspace*{-1em}
    \caption{Minimum \ac{SINR} (i.e., $\min(\Gamma_1,\Gamma_2)$) for different jamming powers for the optimum solution (blue), the \textit{maximum gain} solution (red), the \textit{zero interference} solution (black), and the \textit{vertical orientation} solution (magenta).}
    \label{fig:2}
\end{figure}

We evaluate four distinct cases: (1) Optimum Pose - obtained by numerically optimizing the initial optimization problem~\eqref{eq:2.2a} (blue plot); (2) Zero Interference - using the pose described earlier, we numerically optimize the position according to~\eqref{eq:2.2b} (black plot); (3) Maximum Gain - utilizing the previously mentioned pose, we numerically optimize the position according to~\eqref{eq:2.2c} (red plot); (4) Vertical Orientation - in this case, the antenna orientation vector aligns with gravity, and the position is numerically optimized (magenta plot). This last case represents an under-actuated \ac{MRAV} hovering, where the orientation cannot be controlled independently of the position. From Figure \ref{fig:2}, we make several observations. When the jamming signal is weak, the \textit{maximum gain} solution aligns with the optimum one, and as the jamming signal becomes stronger, the \textit{zero interference} solution becomes optimal. In essence, the suboptimal solutions proposed in this paper coincide with the optimum poses for extreme jamming conditions.

Another observation is the saturation of all solutions when the jamming signal's strength increases. The performance of the \textit{zero interference} solution remains constant because it always nullifies the jamming signal. The optimum solution approaches the \textit{zero interference} solution, eventually nullifying the jamming signal. The \textit{maximum gain} solution keeps maximizing its antenna gain in the links with the legitimate nodes, but starts moving its position so that the null of its antenna is pointed to the jammer. The \textit{vertical orientation} solution moves towards the jammer until the \ac{MRAV} positions itself directly above the jammer, effectively pointing its null towards the malicious node and completely eliminating the jamming signal. These results show the advantages of optimizing the full pose of omnidirectional \acp{MRAV} to establish robust communication in the presence of jamming attacks. In comparing the performance of omnidirectional \acp{MRAV} (blue, red and black plots) to under-actuated ones (magenta plot), significant performance differences are apparent, especially under strong jamming. This technique can also prove valuable when the malicious node ${M}$ acts as an eavesdropper. In such a situation, the objective would shift towards optimizing the pose of the \ac{MRAV} to maximize the secrecy rate. It is worth noting that the frame of the \ac{MRAV} has the potential to modify the radiation pattern of the antenna~\cite{Rizwan2017iAIM}. In forthcoming experiments, we will show how these alterations affect our proposed technique.



\vspace{-0.5em}
\section{Conclusion}
\label{sec:conclusions}
\vspace{-0.5em}

This paper examined the integration of omnidirectional \acp{MRAV} into communication networks, focusing on its pose optimization to mitigate jamming attacks. The framework considered the control of antenna orientation and position to improve the minimal \ac{SINR} of the legitimate network. To tackle the highly nonlinear optimization problem, two suboptimal solutions were proposed that demonstrated their effectiveness in scenarios of low and severe jamming. 
We showed that the pose optimization of omnidirectional \acp{MRAV} can effectively nullify the interference of the jammer,
thus introducing the pose optimization of \acp{MRAV} as a new technique for physical layer security. Future work will consider the effect of the uncertainty in the knowledge of the jammer position. We will also compare, in detail, the performance of the pose optimization against the beamforming technique. In addition, we will study the case where the malicious node is an eavesdropper.


\newpage
\bibliographystyle{IEEEbib}
\bibliography{references_short}

\end{document}